\documentclass[10 pt, conference]{ieeeconf}

\usepackage{acro}
\usepackage[utf8]{inputenc}
\usepackage[ruled,vlined,linesnumbered]{algorithm2e}
\usepackage{amsfonts, amsmath, amssymb}
\usepackage{array}
\usepackage{bm}
\usepackage{cite}
\usepackage{url}
\usepackage{multirow}
\usepackage{booktabs}
\usepackage{graphicx}
\usepackage[switch]{lineno}

\title{Robot Grasping and Manipulation: A Prospective}

\author{
    Claudio Zito$^{1}$
    \\
    $^{1}$ Technology Innovation Institute (TII), Abu Dhabi, United Arab Emirates\\
    Claudio.Zito\atsign tii.ae
}

\newcommand\atsign{@}

\begin{document}
\maketitle




``A simple handshake would give them away''. This is how Anthony Hopkins' fictional character, Dr Robert Ford, summarises a particular flaw of the 2016 science-fiction \emph{Westworld}'s hosts. In the storyline, Westworld is a futuristic theme park and the hosts are autonomous robots engineered to be indistinguishable from the human guests, except for their hands that have not been perfected yet. In another classic science-fiction saga, scientists unlock the secrets of full synthetic intelligence, Skynet, by reverse engineering a futuristic hand.

In both storylines, reality inspires fiction on one crucial point: designing hands and reproducing robust and reliable manipulation actions is one of the biggest challenges in robotics. 

Solving this problem would lead us to a new, improved era of autonomy. A century ago, the third industrial revolution brought robots into the assembly lines, changing our way of working forever. The next revolution has already started by bringing us artificial intelligence (AI) assistants, enhancing our quality of life in our jobs and everyday lives--even combating worldwide pandemics~\cite{danica}.

However, for robots, stepping outside the assembly lines requires a different set of skills that synthetic intelligence still lacks. Such skills are closely entangled with the notion of embodiment. A concept in which intelligence is not only centralised at the brain level but distributed along the body and influenced by its environment.

The field of robot grasping and manipulation has seen an exponential growth of attention from the research community in the last two decades. Researchers have made significant progress in many different areas that feed the perception, planning, and control loop so crucial for these tasks. The renewed interest from the general public, industries, and government agencies has contributed to developing new applications and case scenarios-from simple pick-and-place to handling packages or assembly of mechanical components. Nevertheless, the field has not grown evenly; some challenges received or are still receiving a great deal of attention, while others remain unsolved and unpopular. Yet, others have vanished or changed in their core due to newly available technologies or solution models. 

Reliable grasping and manipulation in real-world applications is still out of reach due to several reasons. At a mechatronic level, simple end-effectors, such as parallel grippers or vacuum cups, eliminate model complexity and redundancy at the cost of strong limitations in the way they can grasp or manipulate the object in-hand. Anthropomorphic end-effectors may provide essential features for manipulation, such as movable thumbs or rolling fingers, but the control complexity and the lack of adequate sensing make these devices still impractical. At an algorithmic level, the robotic manipulation pipeline requires many components that preclude platform independence, and the robustness and resilience of the models are challenged by even minimal changes in the setup or environmental conditions.

Encoding any conceivable setup or condition that a robot may face is not a viable solution. However, there is enough evidence that biological brains do not use this tactic either but work as Bayesian machines whose priors are combinations of model-based and data-driven experience.
Hence, generative models (GMs) such as kernel density estimation (KDE) or deep learning (DL) have become well-established tools in robotics. GMs attempt to learn the true distribution of the data from sampled observations. When faced with previously unseen data, they rely on learned features to find common patterns and compute a set of valid candidate solutions. However, training GMs for robot manipulation needs physical interaction data, which is hard to generate.  

A significant amount of work has been dedicated to robot perception and their ability to deal with unstructured environments, especially extreme ones, e.g.,~\cite{jesus2022}. Since depth cameras and high-precision tactile sensors have become widely accessible, robot perception fast-tracked unlocking potentially game-changer solutions, e.g.~\cite{leonardis}. Nevertheless, any technology has its weaknesses; occlusions and shiny or translucent materials yield an incomplete scene reconstruction, and tactile information may be noisy. Rather than attempt to eliminate the source of uncertainty, robots need to learn how to deal with it. In~\cite{barsky2019}, a deep learning framework is presented for a simulated robot drummer. Audio, video and proprioception sensing data are collected to retrieve the missing information from the other inputs when a modality is faulty. 

Robots should also use perception uncertainty as an indicator to modify their behaviour. High uncertainty should lead to more conservative strategies. Imagine reaching into the fridge to grasp a milk bottle that you can only partially see and how this would affect your reaching strategy. Robots can also achieve this by integrating perception uncertainty into their motion planner, as in~\cite{bib:zito_2016,bib:zito_w2012, bib:zito_w2013, zitoIROS2013,zito_2019}.

Another example of dealing with uncertain perception is provided by the humanoid robots Vito and Boris, respectively developed by the Centro Piaggio at the University of Pisa and the Intelligent Robotic Lab at the University of Birmingham under the European FP7 grant PaCMan~\cite{pacman}. In~\cite{bib:rosales_2018}, the robots outsmart in-hand self-occlusions and vision-driven uncertainty of the object to be manipulated by again combining visual clues and clever tactile exploration of the object surface. 

One of the breakthroughs in grasping and manipulating novel objects was due to contact-based formulations~\cite{bohg2014}. Representation of contacts via establishing a set of relations between (vision-extrapolated) geometrical features on the object's surface and the robot's links have proven to be a powerful solution. Geometrical features are typically extrapolated around the contact points in a paradigm called learning from demonstration (LfD). A teacher is required to present a feasible and robust contact to the robot, but from the geometrical features enough statistic is acquired to learn contact densities in a one-shot fashion as generative contact models~\cite{kopicki2016,arruda2019}. Since it is natural to assume that many objects share many local similarities in terms of geometrical features, these models tend to generalise very well within and across object categories. Task-dependent constraints can be added in the formulation as optimisation procedures, but this requires a good knowledge of the task and, often, ad-hoc solutions. Very recently, a contact-based formulation has also been successfully applied for the first time to the problem of aerial grasping~\cite{zitoaerialgrasping}. Although it should be considered a seminal work, the proposed framework extends the one-shot learning paradigm enabling unmanned aerial vehicles with cable-suspended passive grippers to compute the attach points on novel payloads for aerial transportation with no need for handcrafted task-dependent features.

Generative contact models have also been investigated for tasks beyond grasping. From a young age, humans learn an internal model of how the world works from data-driven experience and curiosity-driven interaction. This internal model plays a crucial role in predicting the outcome of an interaction, even in novel contexts. In~\cite{stuber2018feature,howard2021}, the contact-based formulation enables us to learn an internal model for predicting push motions of previously unseen objects, while in~\cite{zito2012two} a planner uses black-box motion predictors to move objects to the desired configurations. Although the theory behind motion prediction is well-established, the existing methods in the literature are not yet in use in industrial applications. No robotic system can, for example, insert a box of various products onto an over-the-head store shelf by exploiting push operations and the relative contacts and forces generated~\cite{stuber_2019}.     

Of growing interest, it is also the field of physical human-robot interaction (pHRI), where a human operates with a robot to accomplish manipulative tasks~\cite{sheridan}. Remote pHRI is crucial to guarantee the safety of a human operator in extreme and dangerous tasks, such as underwater maintenance, nuclear waste disposal, or rescue missions~\cite{zito2019metrics}. At the same time, wearable robotics, such as prostheses or exoskeletons, have the potential of fully restoring the functionalities of missing body parts or providing super-human capabilities to users~\cite{atzori2014}. Unfortunately, the dexterous capabilities of conventional interfaces for such devices fall a long way short of those we are used to when physically operating with our hands. Additionally, long and tedious training sessions are required for training the users to control the system proficiently. To make more intuitive and accessible interfaces, we need to reliably estimate the user's intention from biological and behavioural clues and map this into appropriate motion commands for the robotic counterpart. In~\cite{veselic2021,zito2019ltv}, an AI assistant for remote teleoperation proactively responds to the user's motion intentions in a predict-then-blend fashion. The system perceives a cluttered scene and, on the fly, predicts candidate grasps for the visible objects and, for each grasp, computes a feasible motion plan. The user commands the robot towards the desired object via a simple interface, e.g., a keyboard or a joystick. The robot attempts to follow the user's wanted motion by latching on to the more similar pre-planned trajectory. 

AI has the potential to revolutionise pHRIs by making them user-friendly and intuitive. For example, activation patterns of muscles can be learned through surface electromyography (sEMG) signals to control prosthetic devices. However, since the biological signal is affected over time due to fatigue, electrode displacement, or sweat, no machine learning system can reliably generate the intended control for a long period of time. Paradigms to overcome this problem have been proposed in the literature, but mostly rely on detecting features that are generally not desired, such  as oscillatory behaviour or high accelerations. It remains unclear whether these models could detect a shift in sEMG patterns, resulting in plausible predictions but producing the wrong hand configuration. In~\cite{heiwolt2019}, a GM-based failure detection was presented that instead spots every instance where the interpretation does not match the user’s intention by incorporating situational context information. Yet, this approach works in a lab where calibrated cameras virtually reconstruct the scene to extrapolate the context. To see this working outside the lab, we would need a more portable technology for 3d reconstruction, which is not ready yet but conceived in the \emph{Mission Impossible}'s smart glasses.

The next decade will be even more exciting for this field. All over the world, roboticists have started moving away from highly-engineered solutions by embracing more flexible and reliable approaches~\cite{sun2021}. At the current state, grasping with imperfect perception is still one of the main issues that slows progress and it will require both research and engineering work~\cite{diluca2012}. Objects with challenging shapes and surfaces can be dealt with but with tailor solutions to known objects, such as model-based 3D reconstructions and by design specific end-effectors.
Hardware and software integration is still tedious and time-consuming, but multiple efforts have been made to alleviate it, e.g. ROS, MoveIT, and we will witness a further increment of such tools in the next years to come. At this pace, it is safe to assume that robust and precise grasping will be consolidated for many different scenarios and applications and we will witness advanced robot pick-and-place in the agricultural industry and delivery services. Beyond pick-and-place tasks, many of the current solutions will fall apart. Grasping for manipulation purposes needs planning while considering task-dependent constrains. Many of these constrains are hard to encode and on-the-fly generation of contacts yields to unreliable solutions even for known objects. This will remain a hard challenge for the next decade on which many researchers will focus their attention. 
In the time of autonomous racing cars, dynamics of a high-speed moving vehicles can be legitimately approximated for optimal control, yet making predictions for manipulation tasks has not received the same attention from the community. Available solutions generate plans to merely avoid collisions assuming quasi-static dynamics by listing an handful of possible outcomes matched against a set of predefined motion strategies. Finite enumeration of states is not an ample strategy for such tasks, and the community will need to investigate new routes for more sustainable solutions.
In-hand manipulation is still at its dawn. Clever designs of tools and end-effectors can achieve specific in-hand manipulation, but without an adequate sensory feedback and clever control strategies this problem remains one of the most challenging tasks a robot can face.  
Finally, in the last decade we have seen an increasing interest towards pHRI and its applications. Exoskeletons and prosthetic devices are getting smarter and a large amount of effort has been, and will be, dedicated to investigate more intuitive interfaces. Augmented and virtual reality technology will play a major role in this field with exciting new immersive solutions.

\bibliographystyle{IEEEtran}
\bibliography{main}
\end{document}